%% file: main.tex
\PassOptionsToPackage{table}{xcolor}
\documentclass[10pt,twocolumn,letterpaper]{article}
\usepackage{multirow}
\usepackage{booktabs}
\usepackage{tabularx}
\usepackage{colortbl}
\usepackage{breqn}
\usepackage[accsupp]{axessibility}  

\usepackage{todonotes}
\setuptodonotes{inline}

\usepackage{cvpr}              

\usepackage[normalem]{ulem}
\useunder{\uline}{\ul}{}


\definecolor{cvprblue}{rgb}{0.21,0.49,0.74}
\usepackage[pagebackref,breaklinks,colorlinks,citecolor=cvprblue]{hyperref}

\newcolumntype{Y}{>{\centering\arraybackslash}X}
\definecolor{lightgray}{rgb}{0.9, 0.9, 0.9}
\newcolumntype{Z}{>{\columncolor{lightgray}}Y}

\def\paperID{28} 
\def\confName{CVPR WAD}
\def\confYear{2024}

\title{Exploring Real World Map Change Generalization of Prior-Informed HD Map Prediction Models}

\author{Samuel M. Bateman$^{\ast ^\dagger}$, Ning Xu$^\ast$, H. Charles Zhao$^\ast$,  Yael Ben Shalom, \\ Vince Gong, Greg Long, Will Maddern\\
Nuro, Inc. \\
{\tt\small \{sbateman, nxu, czhao, ybenshalom, vgong, glong, will\}@nuro.ai}
}

\newcommand\blfootnote[1]{%
  \begingroup
  \renewcommand\thefootnote{}\footnote{#1}%
  \addtocounter{footnote}{-1}%
  \endgroup
}

\begin{document}
\maketitle
\input{sec/0_abstract}    
\input{sec/1_intro}

\input{sec/2_related_work}

\input{sec/3_methodology}

\input{sec/4_experiments}
\input{sec/5_discussion}
\input{sec/6_conclusions}


\end{document}

%% file: sec/0_abstract.tex
\begin{abstract}
Building and maintaining High-Definition (HD) maps represents a large barrier to autonomous vehicle deployment.
This, along with advances in modern online map detection models, has sparked renewed interest in the online mapping problem.
However, effectively predicting online maps at a high enough quality to enable safe, driverless deployments remains a significant challenge.
Recent work on these models proposes training robust online mapping systems using low quality map priors with synthetic perturbations in an attempt to simulate out-of-date HD map priors.
In this paper, we investigate how models trained on these synthetically perturbed map priors generalize to performance on deployment-scale, real world map changes.
We present a large-scale experimental study to determine which synthetic perturbations are most useful in generalizing to real world HD map changes, evaluated using multiple years of real-world autonomous driving data.
We show there is still a substantial sim2real gap between synthetic prior perturbations and observed real-world changes, which limits the utility of current prior-informed HD map prediction models.
\end{abstract}

%% file: sec/1_intro.tex
\section{Introduction}
\label{sec:intro}
\begin{figure}[h]
    \centering
    \begin{minipage}{0.5\textwidth}
        \centering
        \hfill
        \begin{subfigure}[t]{0.5\textwidth}
            \frame{\includegraphics[width=\textwidth,trim={0 0 0 2.5cm}, clip]{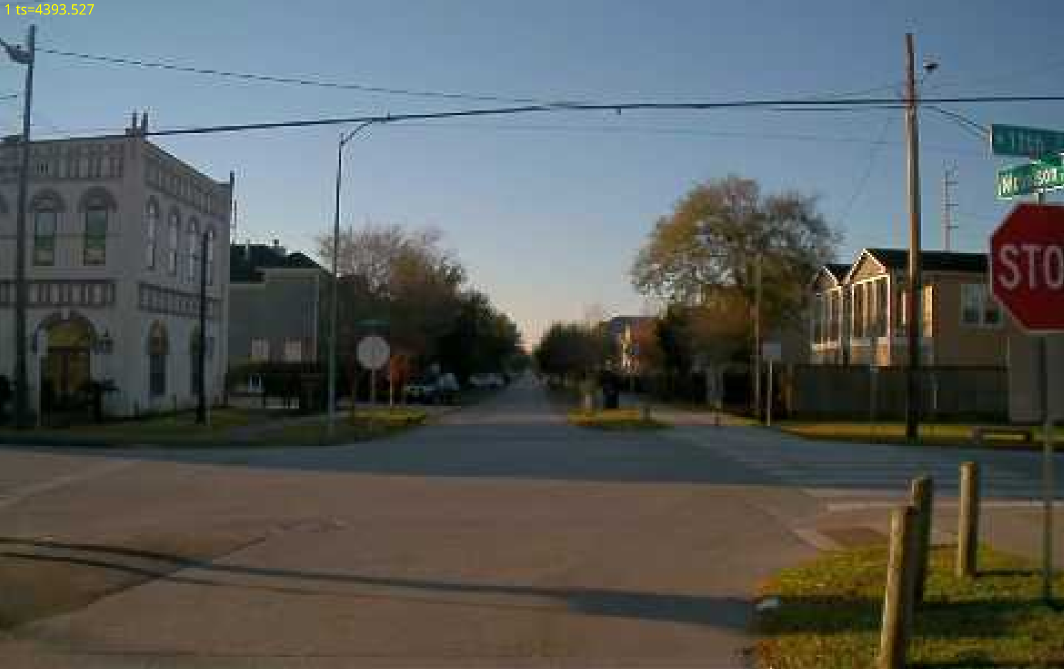}}
            \caption{Camera View Before Change.}
            \label{fig:camera_before_change_cover}
        \end{subfigure}\hfill
        \begin{subfigure}[t]{0.5\textwidth}
            \frame{\includegraphics[width=\textwidth]{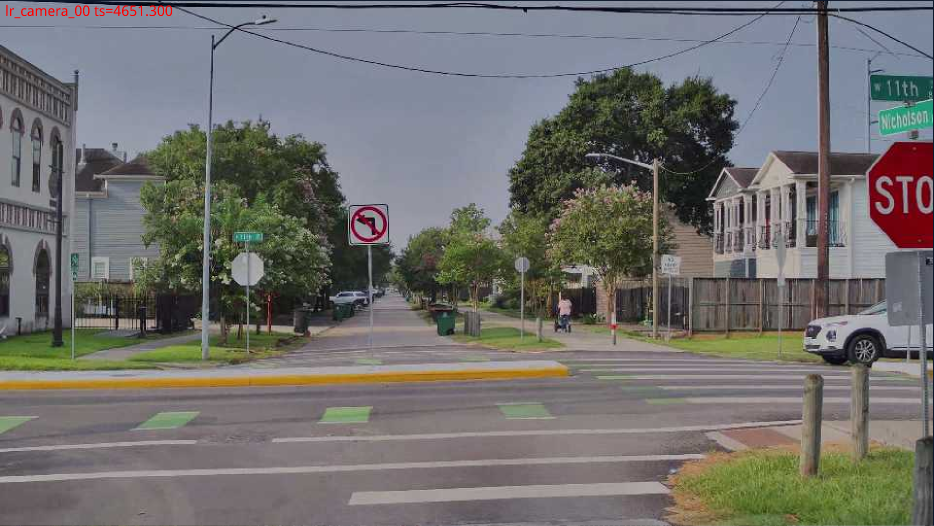}}
            \caption{Camera View After Change.}
            \label{fig:camera_after_change_cover}
        \end{subfigure}\hfill
        \begin{subfigure}[t]{0.5\textwidth}
            \frame{\includegraphics[width=\textwidth]{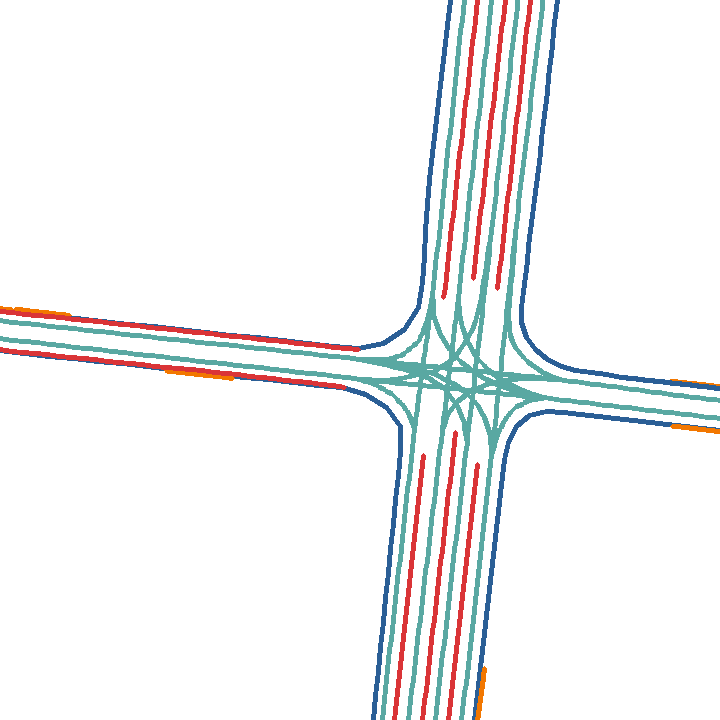}}
            \caption{Example Map Before Change}
            \label{fig:map_before_change_cover}
        \end{subfigure}\hfill
        \begin{subfigure}[t]{0.5\textwidth}
            \frame{\includegraphics[width=\textwidth]{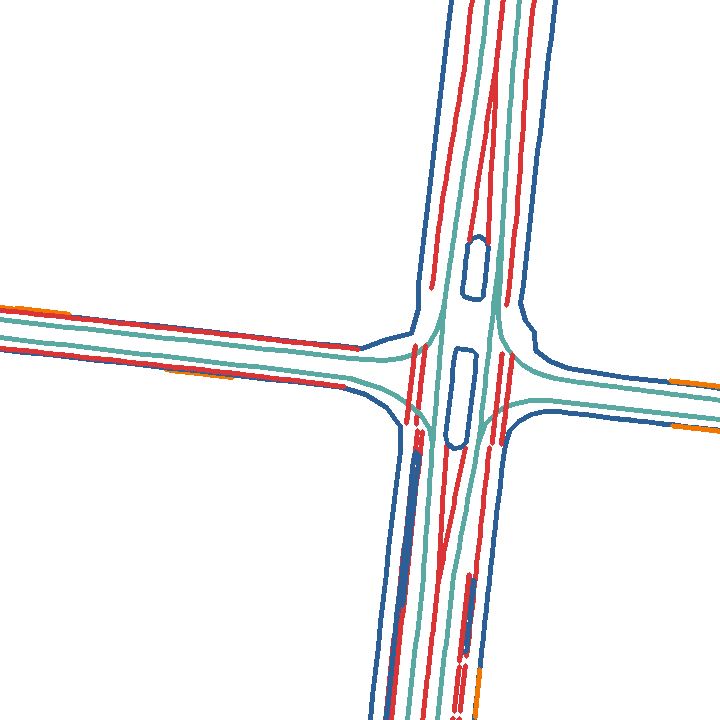}}
            \caption{Example Map After Change}
            \label{fig:map_after_change_cover}
        \end{subfigure}\hfill

        \hfill

    \end{minipage}

    \caption{A real-world map change from an autonomous vehicle dataset. In this paper we investigate which synthetic perturbations applied to a simulated prior map at training time best model a real prior map (left) for training a prior-informed online mapping model to produce the updated map (right), evaluated using a vast collection of real-world changes gathered over multiple years of autonomous vehicle operation.}
    \label{fig:cover_figure}
\end{figure}

\blfootnote{$\ast$ Equal Contribution.}
\blfootnote{$\dagger$ Corresponding Author.}

Large scale autonomous driving has long been a milestone for the robotics community and has been pursued for well over a decade now. 
Perception and mapping systems have formed core components of self driving systems, and mobile robots more generally, enabling comprehension and understanding of the world around them \cite{redmon_you_2016, he_mask_2017}. 
In the field of mobile robotics, the goal of mapping systems at a high level is to predict semi-static geometry and affordances in a scene, i.e. elements of the world that rarely change over time.
These semi-static elements are traditionally encoded with human labeled ``High Definition" (HD) maps built atop a high resolution geometric world reconstruction utilizing many overlapping trajectories.
Reducing this cost within the context of self driving would increase scalability, cost effectiveness, and robust safety. 
However, accurately and consistently constructing online HD maps from sensor data at sufficiently long ranges to facilitate safe, fully autonomous driving has proven challenging (outside of highways and simple road scenarios which have limited topological or geometric complexity).

Despite this, in the last few years, there has been a resurgence of interest in this problem. 
Equipped with modern machine learning techniques \cite{loshchilov_decoupled_2019, cubuk_randaugment_2020}, architectures \cite{vaswani_attention_2017,dosovitskiy_image_2021, carion_end--end_2020}, and open datasets \cite{caesar_nuscenes_2020, wilson_argoverse_2023}, the field has begun to pivot towards this online mapping approach as performance on other previously out of reach perception tasks has become more compelling. 
While exciting, the challenge of predicting highly accurate HD maps at suitable ranges remains, and the majority of recently published research on online mapping degrades in prediction performance with range from the vehicle.

One response to this problem, and one explored by concurrent work in \cite{sun_mind_2023}, is integrating offline, potentially out of date or inaccurate HD Map priors into the prediction of online features. 
This is particularly compelling because low quality, out of date HD maps or sparse, low resolution Standard Definition (SD) maps are often available or cheaply labeled, and maps of roads are generally slow to change over time. 
The result would be a model that only rarely has to perform full online mapping, but most of the time is acting to clean up the discrepancies between the prior and reality. 

A major caveat is that real world examples of meaningful map changes are relatively rare, even from the point of view of large, industrial deployments \cite{lambert_trust_2021}.
To step around this issue, MapEX \cite{sun_mind_2023} instead proposes that one could use synthetic mutations of labels to imitate these map changes over time.
This allows one to generate a virtually limitless number of map changes to train their model to ``fix" the prior, effectively training a change detection/map repair model rather than a full online mapping model.
While this almost certainly would not be trained on the same noise distribution as real changes in the world, one would hope that sufficiently diverse perturbations of the prior map in training would minimize the sim2real transfer gap by acting akin to domain randomization \cite{tobin_domain_2017}, where the real distribution is in the support of the synthetic distribution applied to the labels.
This alternative problem formulation where we have access to a prior could, in theory, significantly simplify the problem for perception systems, and has empirically been shown to outperform existing online-only models \cite{sun_mind_2023}.
However, the broad applicability of such methods is predicated on its ability to effectively transfer from synthetic offline perturbation to real world map change events.

In this paper, we aim to answer the following two questions: 

\begin{itemize}
\item Does training prior-informed online mapping models on synthetic prior mutations \cite{sun_mind_2023} generalize to real world map change examples? 
\item Considering a broad range of prior noise models, how do mixtures of prior noise models applied in training affect the generalization performance of these online mapping models to real world map changes?
\end{itemize}

In addition, we share details of how our model architecture differs from existing online mapping models in the literature to aid in reproducibility of our work.

%% file: sec/2_related_work.tex
\section{Related Work}
\label{sec:related_work}

\subsection{Birds Eye View Perception}
3D perception is a core problem in mobile robotics and computer vision more broadly. The long term trend of 3D perception has been to leverage large, expressive backbones \cite{simonyan_very_2015, he_deep_2015, liu_convnet_2022, dosovitskiy_image_2021, liu_swin_2021, li_bevformer_2022, liu_bevfusion_2022} trained on very large image datasets \cite{russakovsky_imagenet_2015, lin_microsoft_2014, caesar_nuscenes_2020, wilson_argoverse_2023, sun_scalability_2020}, to feed high quality image representations into various heads for specific tasks, such as image classification \cite{krizhevsky_imagenet_2012}, object detection \cite{redmon_you_2016, liu_ssd_2016, he_mask_2017, carion_end--end_2020}, semantic segmentation \cite{badrinarayanan_segnet_2015, xiao_unified_2018}, keypoint estimation \cite{he_mask_2017}, and more. This trend has only strengthened with the rise of very large models pretrained on largely unsupervised objectives with internet scale data both within computer vision \cite{oquab_dinov2_2024} and outside of it \cite{brown_language_2020}.

One direction that has advanced considerably in the last few years is the representation space used for 3D perception.
Early 3D object detection approaches focused on a couple key approaches: one being detecting and tracking in 2D image space and reprojecting model outputs to 3D space using geometric information \cite{pang_clocs_2020}, and the other being early fusion through methods like \cite{vora_pointpainting_2020}.
With the rise of the use of LiDAR in mobile robotics perception tasks, efforts were made to develop better data representations and encoders for 3D object detection with LiDAR \cite{qi_pointnet_2017, zhou_voxelnet_2018, lang_pointpillars_2019, qi_frustum_2018}.
Because of the natural 3D geometry of LiDAR data and the approximately 2.5D worlds that mobile ground robots generally perceive (i.e., generally much fewer detections coinciding along the z direction than in the x and y directions), the aforementioned publications cumulatively proposed laying out LiDAR information as features in a voxel grid represented as a top down feature image.
This topdown representation, known as the Birds Eye View (BEV) representation, has attracted a large amount of attention in perception for mobile robotics over the past few years.
In particular, a huge number of contributions have sought to develop BEV representations of other sensors as well so that they can be simply merged into a single unified feature representation. 
For example, much of this research focus has been on developing expressive BEV feature backbones utilizing both camera and LiDAR data \cite{philion_lift_2020, liu_bevfusion_2022, li_bevformer_2022, zhang_occformer_2023}, which then can be consumed by downstream tasks through a relatively simple, single BEV image interface.

\subsection{Online HD Map Construction}
One specific downstream task which has been explored on top of these BEV models for the task of autonomous driving is online mapping.
Early efforts in online mapping primarily focused on doing semantic segmentation from the perspective view of a camera \cite{yu_bdd100k_2020}, or as a semantic segmentation problem utilizing a BEV representation \cite{li_hdmapnet_2022}. 
However, these methods struggle with two key problems: real world roads often have complex topologies and instance-level traits which are difficult to represent accurately with semantic segmentation, and many downstream behavior planning models (e.g. \cite{nayakanti_wayformer_2023}, \cite{jiang_vad_2023}) consume vectorized map representations.
Indeed, part of the reason why HD maps are generally labeled as vectorized features is that this removes much of the ambiguity regarding topology and overlapping elements and affordances of roads.
Thus, it would be ideal to directly predict vectorized features.
Earlier work identified this requirement, and reframed the polyline prediction problem in a few different ways, i.e. autoregressive transformer approaches \cite{liu_vectormapnet_2023} or classical, heuristic post processing of semantic segmentation \cite{li_hdmapnet_2022}.
A recent line of work \cite{liao_maptr_2023, liao_maptrv2_2023} similarly reframed the problem again by making the connection that vectorized online mapping can be represented as an unordered set detection problem, with a similar problem structure as the one-to-one bijective object detection transformer described in \cite{carion_end--end_2020}, and reported compelling empirical performance with this approach.
This has sparked a flurry of renewed interest in the online mapping problem \cite{liao_maptrv2_2023, lilja_localization_2023, liu_leveraging_2024}. The work most related to ours is the concurrent work of MapEX \cite{sun_mind_2023} which proposes to add low quality HD map priors as inputs to MapTR \cite{liao_maptr_2023} to improve detection performance despite occlusion or changes in the underlying map, attempting to solve the well known map change problem.
This provides an alternative avenue for solving the change detection problem: rather than maintaining an up to date map, one can instead train a model to leverage both sensor data and out of date HD map data to reconstruct an accurate, live representation of the world around it, enabling lifelong deployment of mobile robots with significantly lower HD map maintenance expense.

%% file: sec/3_methodology.tex
\section{Methodology}
\label{sec:methodology}
\subsection{Map Detection Head}
To predict vectorized map features, we follow in the footsteps of MapTR \cite{liao_maptr_2023, liao_maptrv2_2023} with a Deformable DETR \cite{zhu_deformable_2021} based polyline detection model, with one query per control point where this query is the sum of a per polyline embedding and a per point embedding as in \cite{liao_maptr_2023}.
In the following sections, we will primarily describe places where our methodology diverges from existing literature to aid in reproducibility of our results.

\begin{figure*}[htpb]
  \centering
  \includegraphics[width=\linewidth]{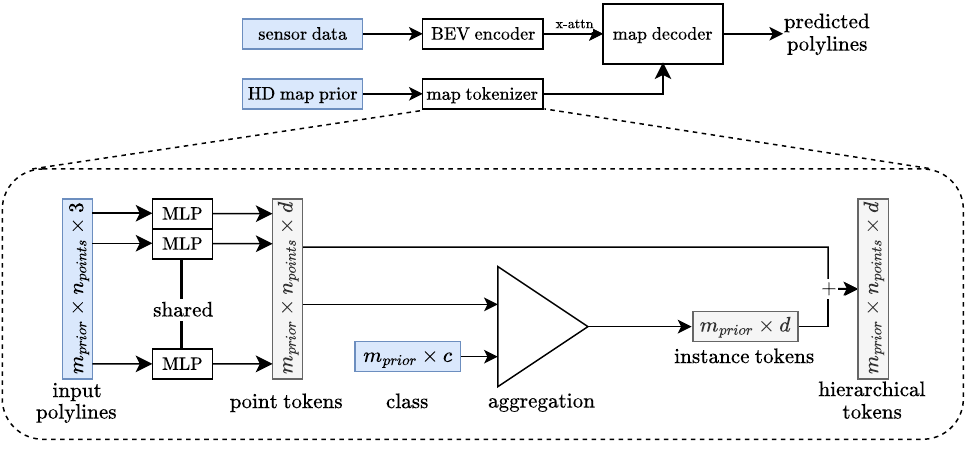}
  \caption{Overall architecture of our model, heavily influenced by \cite{liao_maptr_2023,sun_mind_2023}. See these references for more details on implementation.}
  \label{fig:arch}
\end{figure*}
\subsection{Single Stage Bipartite Matching Loss}
We forgo the two-stage hierarchical matching loss from MapTR \cite{liao_maptr_2023} and instead utilize a simpler one stage bipartite loss by computing a pairwise loss matrix before computing the bipartite matching itself.
To do this, we first compute a set of loss tensors to model the positional error and the permutation symmetries introduced in \cite{liao_maptr_2023}.
We will use an indexed matrix notation to define the constructed tensors used to compute the losses, where $\hat{x} \in \mathbb{R}^{m_{pred } \times n_{points} \times p_{dim}}$ is the prediction from the decoder transformer, where $m_{pred}$ is the maximum number of predictions, $n_{points}$ is the number of control points in every polyline, and $p_{dim}$ is the dimensionality of each control point.
Similarly, $x \in \mathbb{R}^{m_{gt} \times n_{points} \times p_{dim}}$ is the ground truth polyline tensor padded with no-object classes to always have $m_{gt}$ objects to match to, where $m_{gt}$ is the max number of ground truth labels for a given frame.
Since $m_{pred}$ is fixed, we set $m_{pred} = m_{gt}$.
Following this, we construct our loss matrices in the general form
\begin{equation}
  \mathcal{L}_{p2p_{c}} = \Bigr[ I_{c}(j) \min_{P_{a} \in Q_c} \sum_{k, l} | P_{a} \cdot \hat{x}_{i, k, l} - x_{j,k,l} | \Bigr]_{i, j}
  \label{eq:general_point2point}
\end{equation}
where 
\begin{multline}
Q_c = \{P_a | P_{a} \in \mathbb{M}^{m_{pred} \times m_{pred}}, \\
              P_a \in \text{valid permutations for class c}\}
  \label{eq:permutation_set}
\end{multline}
is the set of valid permutation matrices for a given invariance class $c \in \{$polygon, undirected polyline, directed polyline$\}$ and 
\begin{equation}
I_c(j) = \begin{cases}
1 & \text{Invariance Class}(j) = c \\
0 & otherwise
\end{cases}
  \label{eq:invariance_mask}
\end{equation}
is a masking function which sets the $j$th ground truth label loss row to all $0$ if the ground truth loss is not of that invariance class.
As in \cite{liao_maptr_2023}, the valid permutations of polygons represent the set of all shift permutation maps with the polyline indexed in both directions (clockwise and counter-clockwise).
Similarly, the valid set of permutations for undirected polylines is only swapping directions of the polyline and is simply the identity permutation for directed polylines.
Using an actual permutation matrix would be computationally expensive, but these can be implemented inexpensively using \texttt{tf.roll}, \texttt{tf.reverse}, and \texttt{tf.tile} operations in Tensorflow \cite{abadi_tensorflow_2016} or similar operations available in most any comparable differentiable array computation library \cite{paszke_pytorch_2019, bradbury_jax_2018}.
Note that we optionally also add a scaled pairwise cosine similarity loss from \cite{liao_maptr_2023} to this as well, though our experience suggests that weighting cosine similarity much lower than the pointwise positional loss helps with convergence when using this single step training objective.

With these 3 loss matrices, we can construct a single stage point2point matrix loss as:
\begin{equation}
\mathcal{L}_{p2p}(\hat{x}, x) = \mathcal{L}_{p2p_{u}}(\hat{x}, x) + \mathcal{L}_{p2p_{d}}(\hat{x}, x) + \mathcal{L}_{p2p_{p}} (\hat{x}, x)
  \label{eq:point2point_loss}
\end{equation}
where $\mathcal{L}_{p2p_{u}},\mathcal{L}_{p2p_{d}}$, and $\mathcal{L}_{p2p_{p}}$ refers to the resulting pairwise loss matricies for each invariance class, those being undirected polylines, directed polylines, and polygons respectively.
Note that the resulting per batch polyline localization error matrix includes the pairwise positional error between every ground truth and predicted polyline, where each pairwise positional error is computed with the loss minimizing prediction permutation which is valid for a given pair's label.
This matrix is identically 0 for any no-object ground truth class row, and correctly masks invalid permutations from matching with any given label as defined by the label's invariance class.
We can then utilize a pairwise focal classification loss \cite{lin_focal_2020} matrix $\mathcal{L}_{focal}$ of a similar construction and sum them into a single loss matrix 
\begin{equation}
  \mathcal{L}_{pairwise}(\hat{x}, x)  = w_{c} \cdot \mathcal{L}_{focal}(\hat{x}, x)  + w_{p} \cdot \mathcal{L}_{point2point}(\hat{x}, x) 
  \label{eq:matrix_loss} 
\end{equation}
where $w_c, w_p \in \mathbb{R}$ are weighting terms to each respective loss type.
With this full pairwise loss matrix, we can solve for the minimum loss label/prediction assignment using the Hungarian Algorithm \cite{kuhn_hungarian_1955} and directly sum up the resulting optimally matched losses rather than doing a hierarchical matching as in \cite{liao_maptr_2023}. 
This should result in a similar final cost function, but performing Hungarian matching directly on the final losses ensures there is no divergence of objectives between point-wise and polyline-wise convergence and simplifies training code.

\subsection{Map Tokenizer and Prior Integration}
To incorporate an HD map prior, we use a map tokenizer similar to \cite{sun_mind_2023}.
The map tokenizer is a lightweight learned module that converts an unordered set of polylines into tokens, the language of transformers (see \cref{fig:arch}).
We want these tokens to encode as much useful information from the prior as possible, including both point-level and polyline-level information.
We take inspiration from the idea of hierarchical queries in MapTR \cite{liao_maptr_2023}, and we set the tokens equal to the sum of a point token and an aggregated polyline token derived from the point tokens.
The point tokens are generated using a Multi-Layer Perceptron (MLP) over the point coordinates, where this MLP is shared across all points.
The polyline tokens are generated by max pooling over point tokens, concatenating this with a one-hot class vector, and passing through another MLP which is shared across all polylines.
This weight-sharing scheme preserves permutation invariance among polylines, and the max pooling is a lightweight way to aggregate information.

One important case to handle is when $m_{prior} < m_{pred}$, where $m_{prior}$ is the number of prior polylines, which is almost always the case. 
If we just naively add padding to the end of the prior up to $m_{pred}$, then some of the positional encodings in the transformer decoder will always be associated with a prior while some will almost never be associated with a prior, which could cause undesirable biases. 
To mitigate this, we shuffle the prior tokens after adding padding so that padding is effectively inserted randomly.

Once we have converted the prior into tokens, we still need a way to consume these tokens.
One approach is to add an extra cross-attention step to the decoder layer that attends to these tokens.
In this formulation, the prior is simply another modality to attend to, in addition to the BEV embedding.
However, we found that this approach failed basic overfitting experiments.
We suspect that cross-attention does not provide enough bandwidth for the model to incorporate the prior as strongly as it should, especially with a limited number of decoder layers.

Another approach is to directly replace the fixed hierarchical queries in MapTR \cite{liao_maptr_2023} with these prior tokens.
This formulation has a nice intuitive interpretation---the prior tokens provide the initial estimates, which are then refined through several decoder layers by attending to the BEV embedding to come up with a posterior.
We found that, when tested against synthetic map perturbations, this approach works better and has more stable training, which is consistent with the approach in \cite{sun_mind_2023}, and thus is the approach used in all of our results.

\begin{figure*}[t!]
    \centering
    \begin{subfigure}[t]{0.24\textwidth}
        \frame{\includegraphics[width=\textwidth]{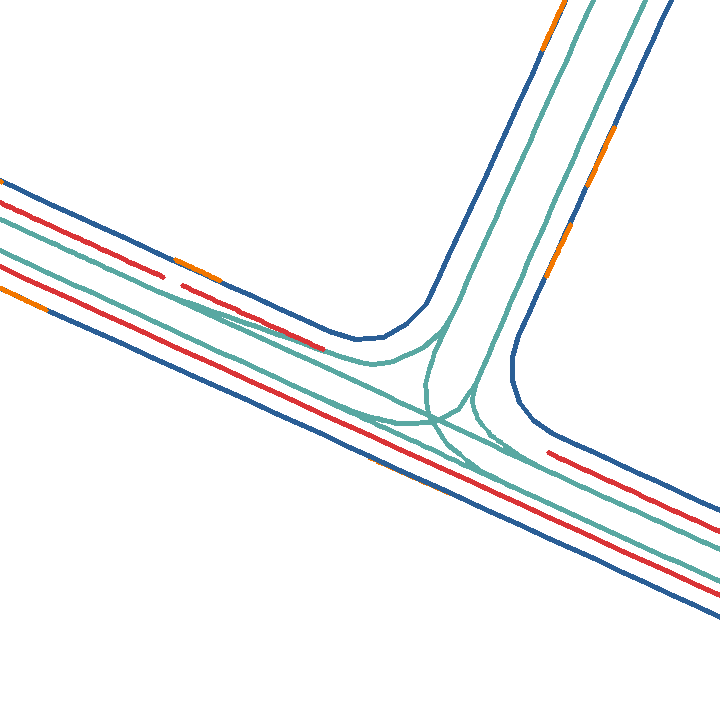}}
        \caption{No perturbations.}
        \label{fig:no_perturb}
    \end{subfigure}\hfill
    \begin{subfigure}[t]{0.24\textwidth}
        \frame{\includegraphics[width=\textwidth]{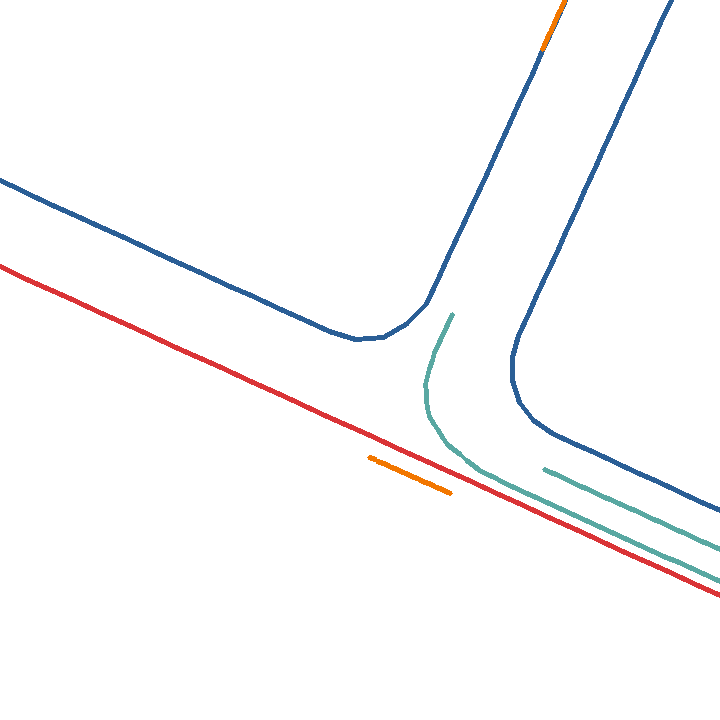}}
        \caption{Feature dropout.}
        \label{fig:dropout}
    \end{subfigure}\hfill
    \begin{subfigure}[t]{0.24\textwidth}
        \frame{\includegraphics[width=\textwidth]{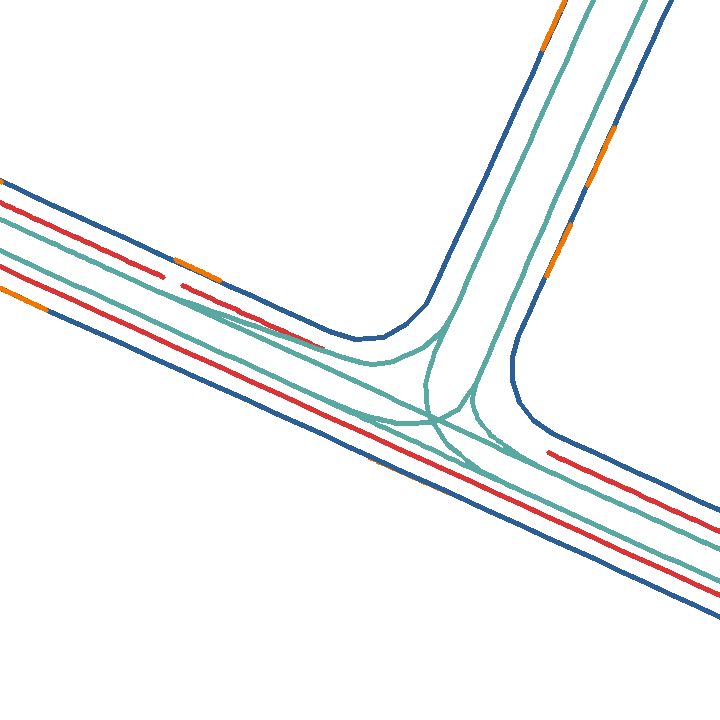}}
        \caption{Feature duplication.}
        \label{fig:duplication}
    \end{subfigure}\hfill
    \begin{subfigure}[t]{0.24\textwidth}
        \frame{\includegraphics[width=\textwidth]{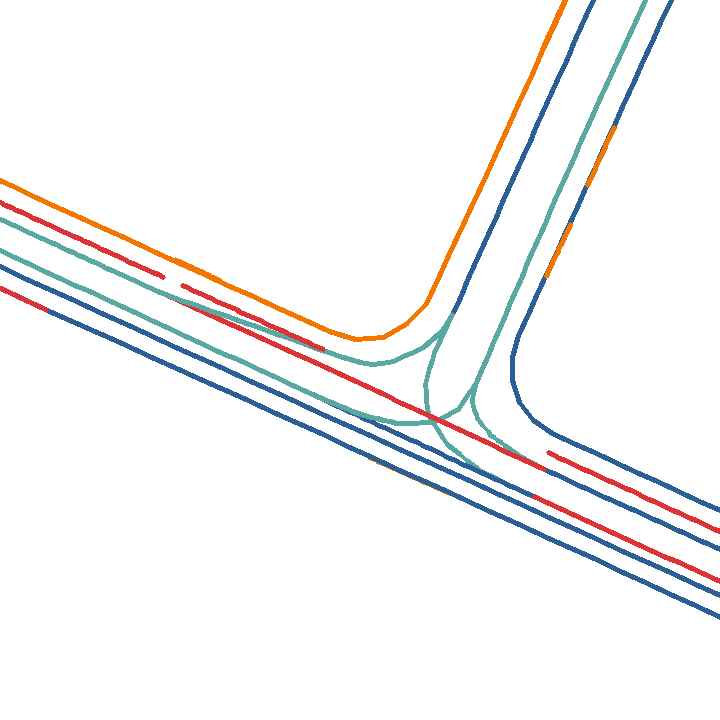}}
        \caption{Wrong class.}
        \label{fig:wrong_class}
    \end{subfigure}
    
    \begin{subfigure}[t]{0.24\textwidth}
        \frame{\includegraphics[width=\textwidth]{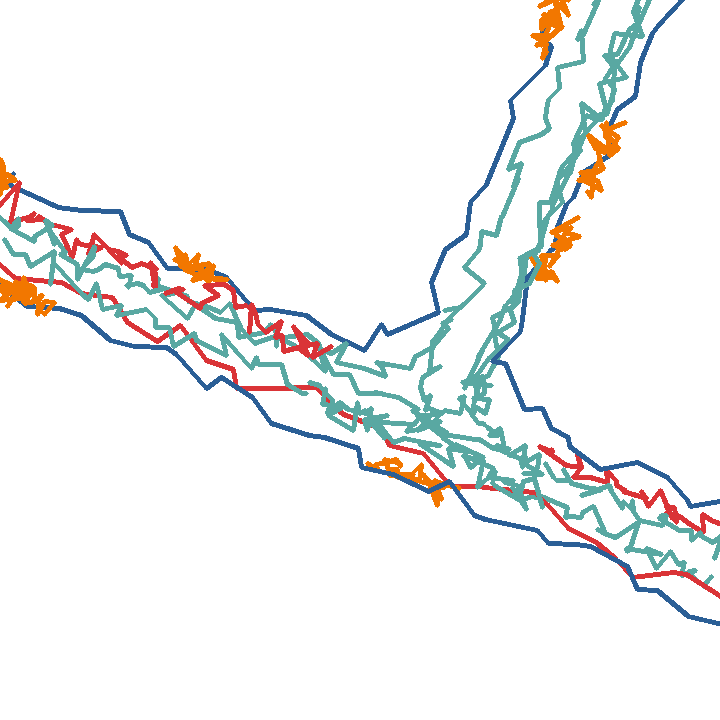}}
        \caption{Control point perturbation.}
        \label{fig:control_point}
    \end{subfigure}\hfill
    \begin{subfigure}[t]{0.24\textwidth}
        \frame{\includegraphics[width=\textwidth]{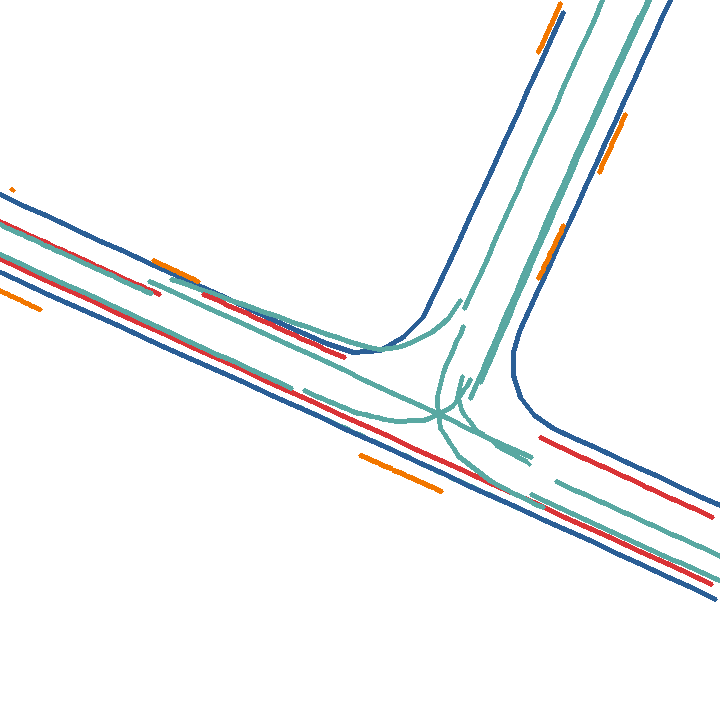}}
        \caption{Feature location perturbation.}
        \label{fig:feature_loc}
    \end{subfigure}\hfill
    \begin{subfigure}[t]{0.24\textwidth}
        \frame{\includegraphics[width=\textwidth]{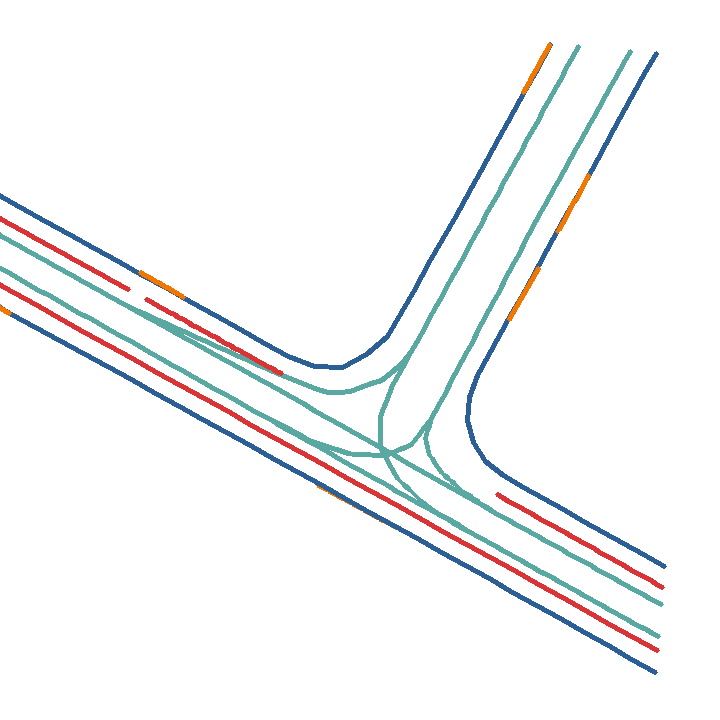}}
        \caption{Global rotation and shift.}
        \label{fig:loc}
    \end{subfigure}\hfill
    \begin{subfigure}[t]{0.24\textwidth}
        \frame{\includegraphics[width=\textwidth]{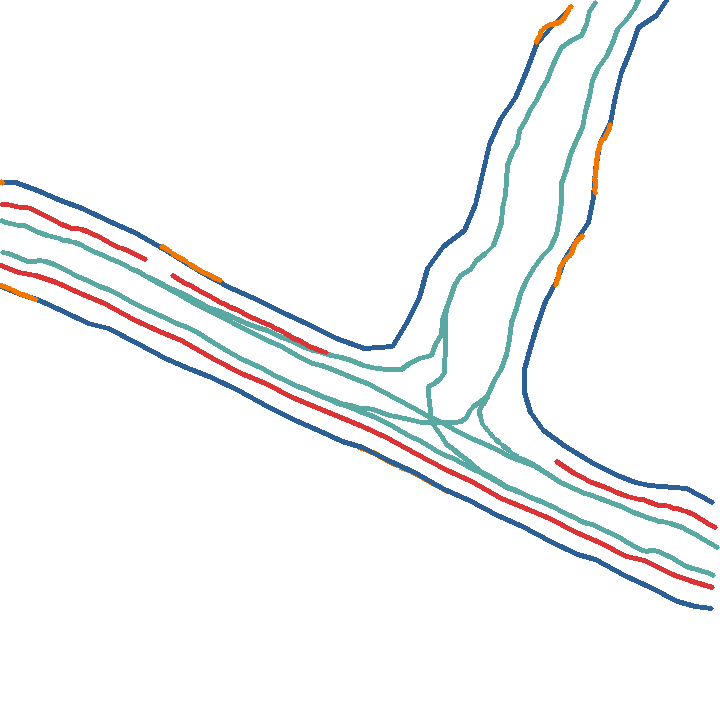}}
        \caption{Perlin noise.}
        \label{fig:perlin}
    \end{subfigure}

    \caption{Types of perturbations. Note that \cref{fig:duplication} looks the same as \cref{fig:no_perturb} since features are exactly duplicated.}
    \label{fig:perturbations}
\end{figure*}

\subsection{Types of Perturbations}
We implement a number of synthetic map prior perturbation types, expanding on the experiments from \cite{sun_mind_2023}.
These can be roughly classified as \textit{discrete mutations} which change the number or types of polyline features, or \textit{continuous, warping mutations} which change the position or shape of the features (see \cref{fig:perturbations}).
Ultimately the goal of all these mutations is to prevent the trained model from simply passing through the vectorized polyline prior features while completely ignoring sensor observations.
\subsubsection{Discrete Mutations}
The primary goal of discrete mutations is to capture map changes to a scene which cannot be fully characterized simply by warping the geometries of the underlying features, and are more similar to a discete Bernoulli distribution of something about a map feature that has or hasn't changed, such as the class of a feature or the number of features in a scene.
\begin{itemize}
    \item \textit{Feature dropout} (\cref{fig:dropout}) -- We use a full feature dropout mutation to model large scene changes or incomplete labels. We drop out each feature in the scene with a Bernoulli distribution with equal probability across each feature.
    \item \textit{Feature duplication} (\cref{fig:duplication}) -- We use a feature duplication mutation to model accidental label duplication, as well as to model large scene changes when mixed with warping perturbations which will warp each duplicated feature in a different way. This mutation works by again using a Bernoulli trial for each existing polyline feature, and truncating the newly added features to a maximum of $m_{pred}$ features.
    \item \textit{Wrong class} (\cref{fig:wrong_class}) -- We use a wrong class mutation which, again by Bernoulli trial, mutates each polyline's class with some probability to a random other class. This has similar goals as previous mutations of decreasing model reliance on the prior, but is the only mutation which corrupts class information, modeling mislabeled features as well as map change events such as repainting of lines or the addition of new lanes.
\end{itemize}

\subsubsection{Continuous, Warping Mutations}
To complement our discrete mutations of category and cardinality, we also introduce a number of continuous, geometry warping mutations.
Note that each of these are parameterized by a standard deviation which scales the variance of the resulting noise.
\begin{itemize}
    \item \textit{Control point perturbation} (\cref{fig:control_point}) -- This is a simple per-control-point zero-mean Gaussian shift to help ensure the model cannot simply pass through the prior as an identity function.
    \item \textit{Feature location perturbation} (\cref{fig:feature_loc}) -- We hypothesize that the model may relatively easily overcome control point perturbation by simply smoothing out the prior rather than attending to sensor data, so we also use a zero-mean Gaussian per-feature location perturbation.
    \item \textit{Global rotation and shift} (\cref{fig:loc}) -- We apply Gaussian perturbations to global yaw and position to simulate robot localization errors. We also hypothesize that even per-feature location perturbation may be relatively easily overcome by the model through self-attention without attending to sensor data, but this localization mutation can only be inverted by leveraging sensor data.
    \item \textit{Perlin warp} (\cref{fig:perlin}) -- We generate two 2D Perlin noise \cite{perlin_improving_2002} images utilizing fractional Brownian motion \cite{kenneth_falconer_fractal_2014}, one each corresponding to warping x and y coordinates, and then renormalize the resulting noise distribution to be zero mean. We then sample from this image at the coordinates of the control points of the polylines to get structured noise. We hypothesize that such correlated warping will make it harder for the model to learn to simply denoise the zero-mean noise added to individual features, better simulating how curbs, lanes, and lane boundaries may all be moved together during a major, real map change event.
\end{itemize}

%% file: sec/4_experiments.tex
\section{Experiments}
\label{sec:experiments}

\begin{figure*}[]
    \vspace{10mm}
    \centering
    \begin{subfigure}[m]{0.24\textwidth}
        \frame{\includegraphics[width=\textwidth]{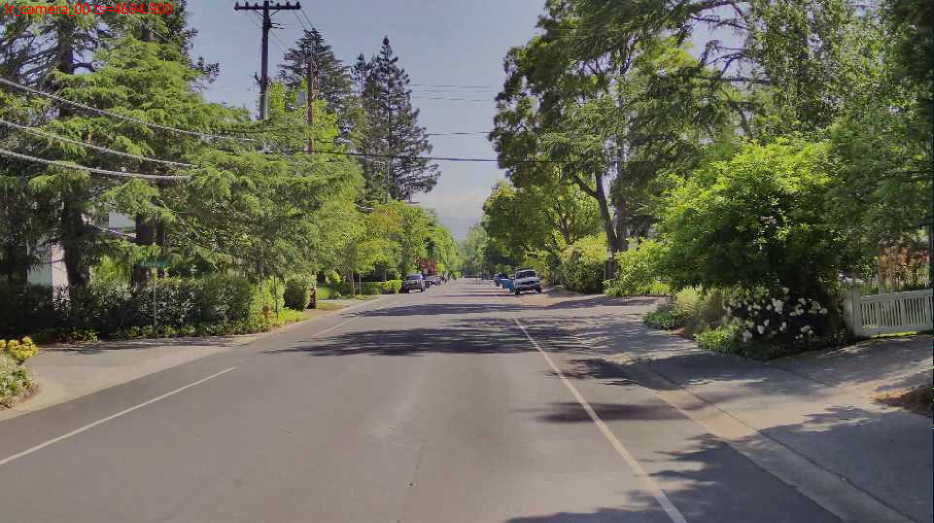}}
        \caption{New Driveway Geometry Scene}
        \label{fig:}
    \end{subfigure}\hfill
    \begin{subfigure}[m]{0.24\textwidth}
        \frame{\includegraphics[width=\textwidth]{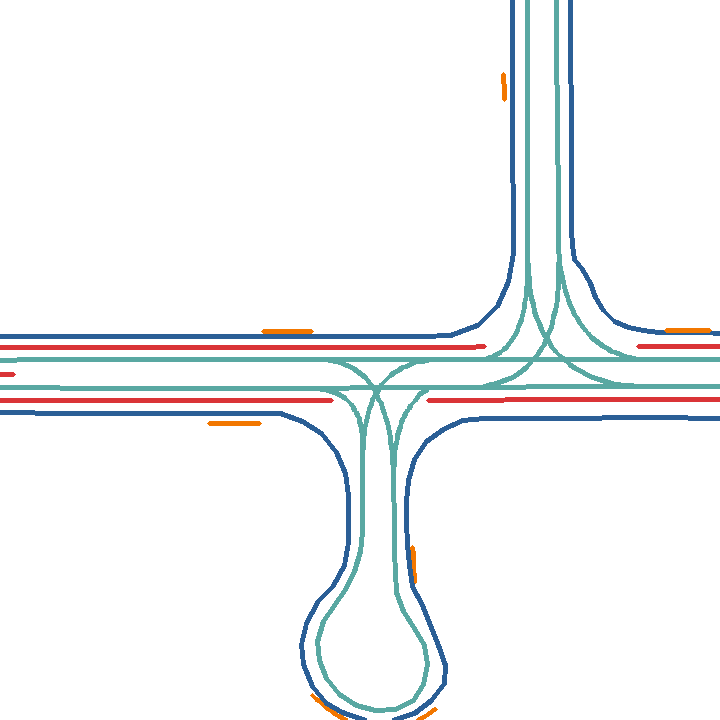}}
        \caption{Outdated Map Prior}
        \label{fig:}
    \end{subfigure}\hfill
    \begin{subfigure}[m]{0.24\textwidth}
        \frame{\includegraphics[width=\textwidth]{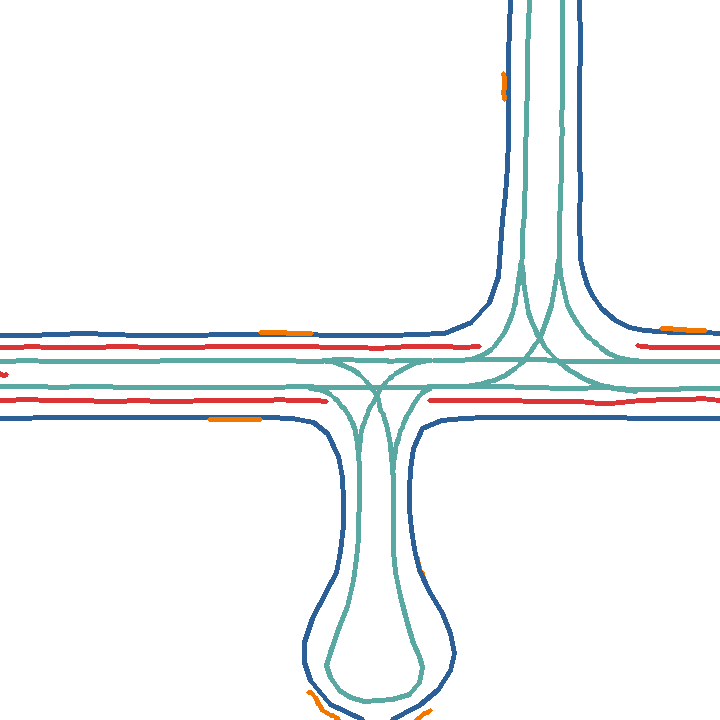}}
        \caption{Prior-Informed Prediction}
        \label{fig:}
    \end{subfigure}\hfill
    \begin{subfigure}[m]{0.24\textwidth}
        \frame{\includegraphics[width=\textwidth]{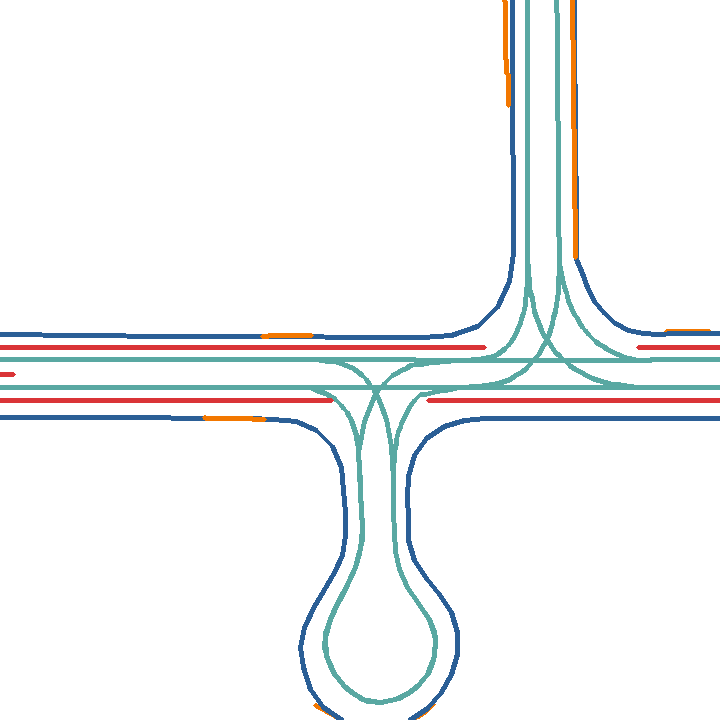}}
        \caption{Ground Truth}
        \label{fig:}
    \end{subfigure}

    \begin{subfigure}[m]{0.24\textwidth}
        \frame{\includegraphics[width=\textwidth]{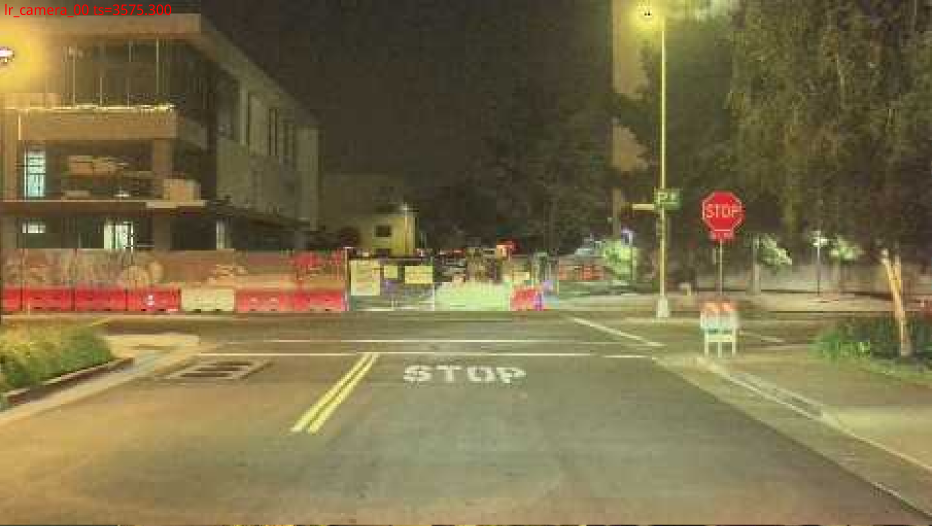}}
        \caption{New Curb Geometry Scene}
        \label{fig:}
    \end{subfigure}\hfill
    \begin{subfigure}[m]{0.24\textwidth}
        \frame{\includegraphics[width=\textwidth]{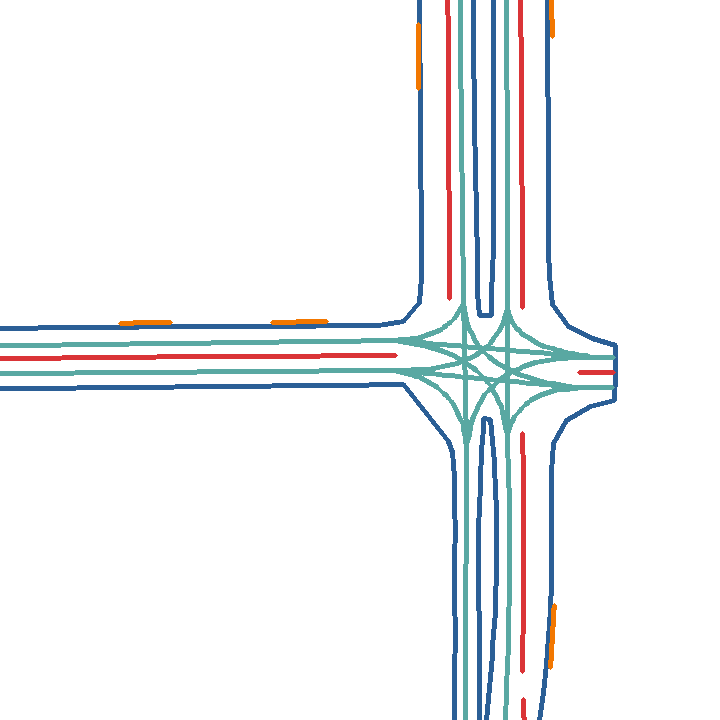}}
        \caption{Outdated Map Prior}
        \label{fig:}
    \end{subfigure}\hfill
    \begin{subfigure}[m]{0.24\textwidth}
        \frame{\includegraphics[width=\textwidth]{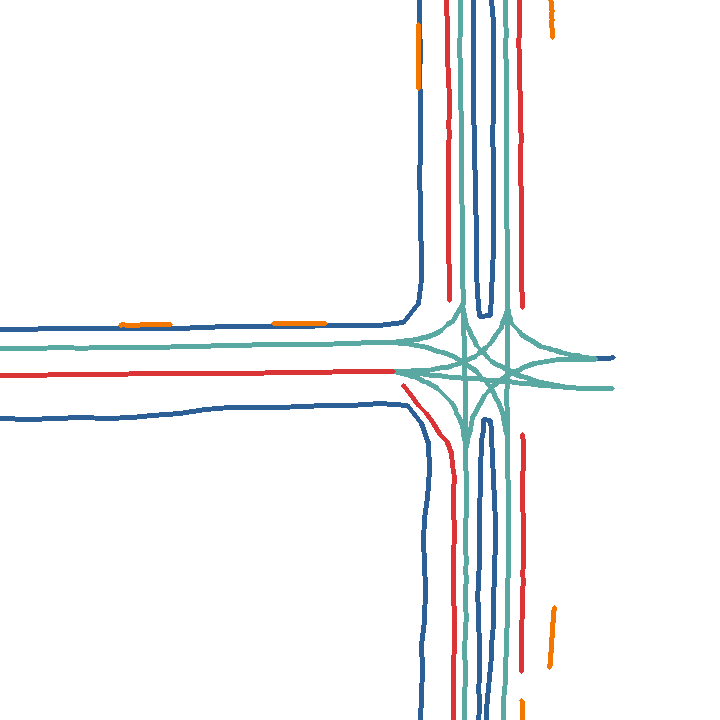}}
        \caption{Prior-Informed Prediction}
        \label{fig:}
    \end{subfigure}\hfill
    \begin{subfigure}[m]{0.24\textwidth}
        \frame{\includegraphics[width=\textwidth]{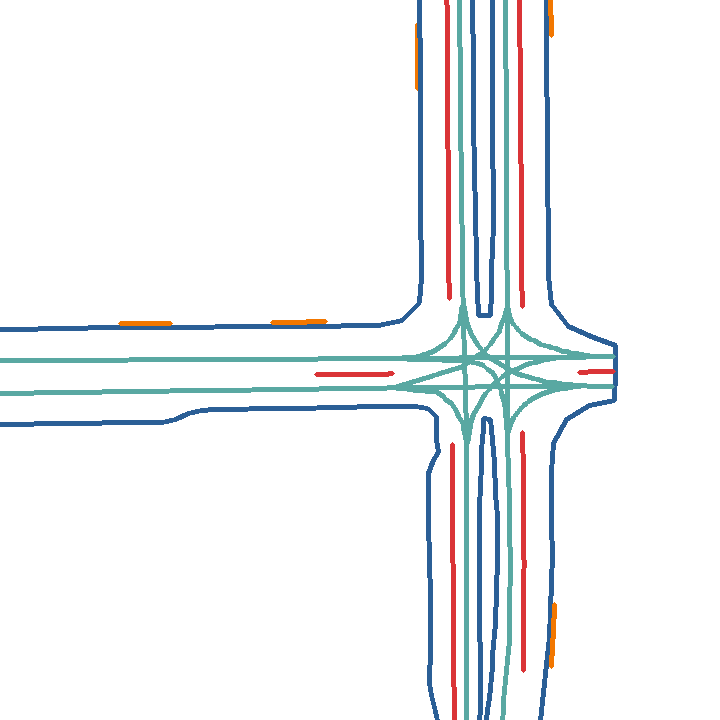}}
        \caption{Ground Truth}
        \label{fig:}
    \end{subfigure}
    
    \begin{subfigure}[m]{0.24\textwidth}
        \frame{\includegraphics[width=\textwidth]{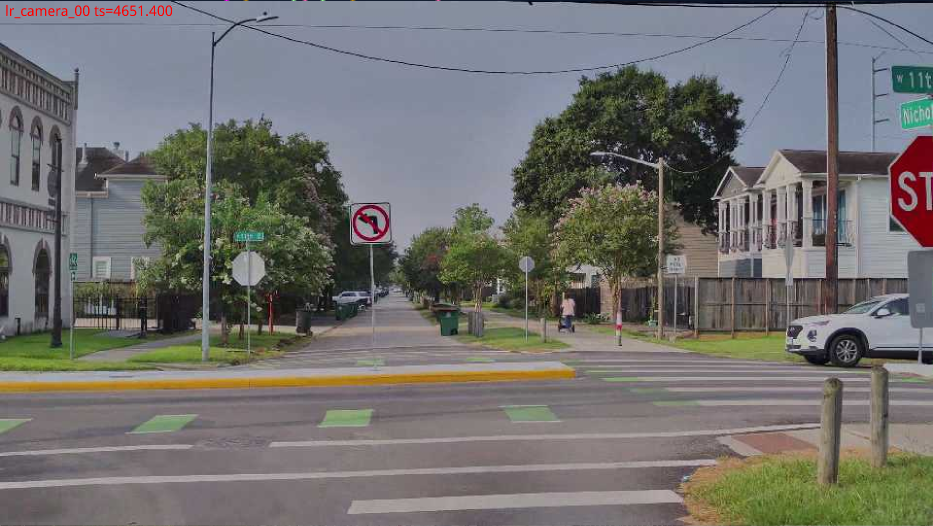}}
        \caption{New Median Intersection Scene}
        \label{fig:}
    \end{subfigure}\hfill
    \begin{subfigure}[m]{0.24\textwidth}
        \frame{\includegraphics[width=\textwidth]{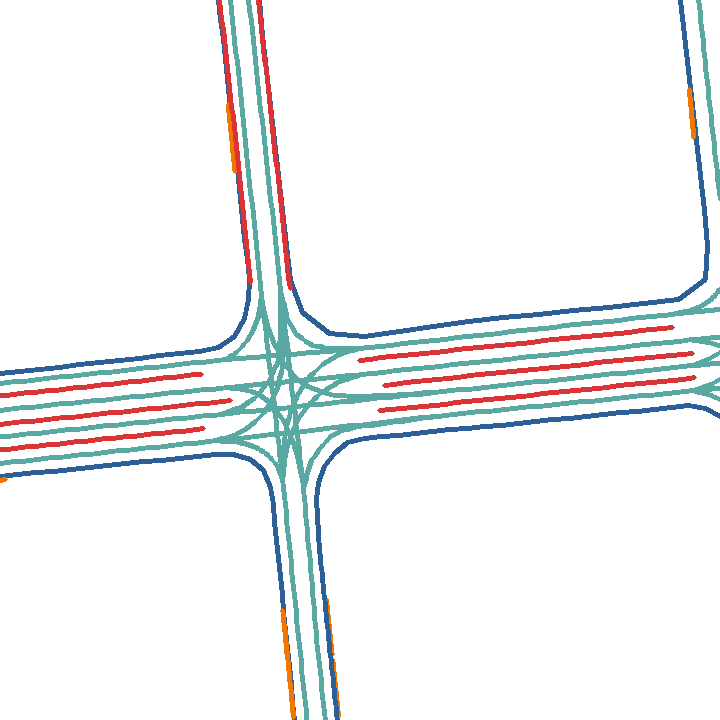}}
        \caption{Outdated Map Prior}
        \label{fig:}
    \end{subfigure}\hfill
    \begin{subfigure}[m]{0.24\textwidth}
        \frame{\includegraphics[width=\textwidth]{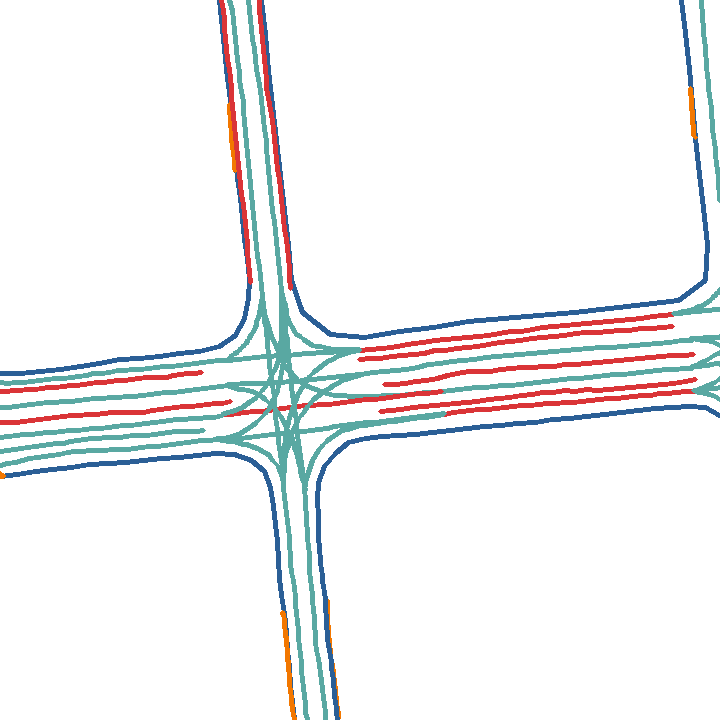}}
        \caption{Prior-Informed Prediction}
        \label{fig:}
    \end{subfigure}\hfill
    \begin{subfigure}[m]{0.24\textwidth}
        \frame{\includegraphics[width=\textwidth]{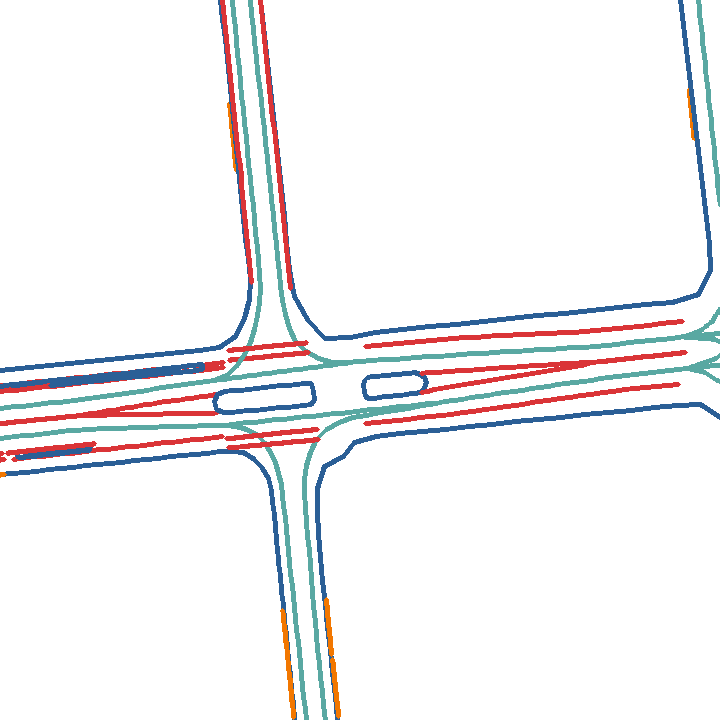}}
        \caption{Ground Truth}
        \label{fig:}
    \end{subfigure}

    \begin{subfigure}[m]{0.24\textwidth}
        \frame{\includegraphics[width=\textwidth]{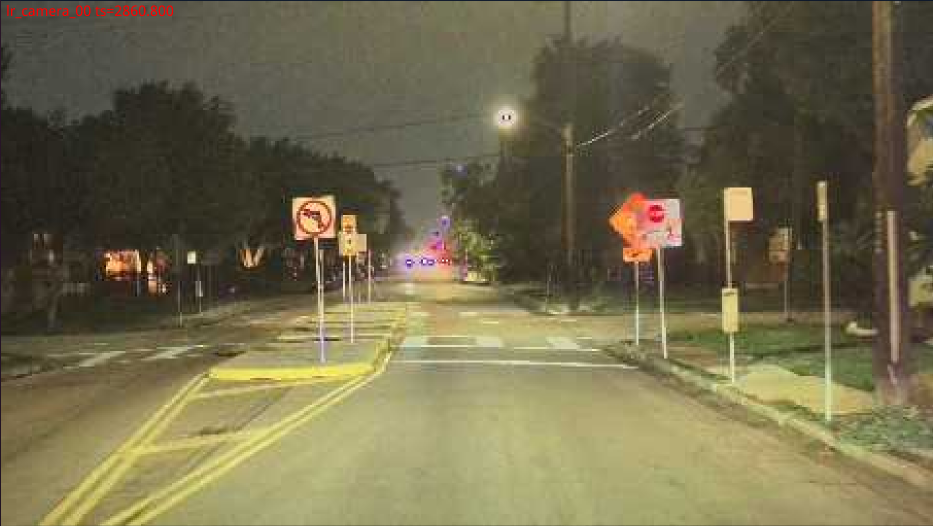}}
        \caption{New Road Construction Scene}
        \label{fig:}
    \end{subfigure}\hfill
    \begin{subfigure}[m]{0.24\textwidth}
        \frame{\includegraphics[width=\textwidth]{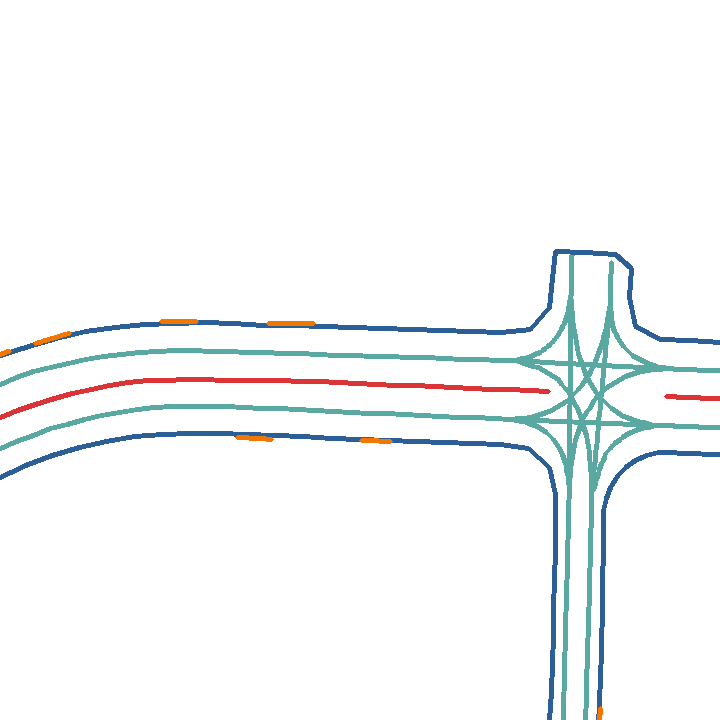}}
        \caption{Outdated Map Prior}
        \label{fig:}
    \end{subfigure}\hfill
    \begin{subfigure}[m]{0.24\textwidth}
        \frame{\includegraphics[width=\textwidth]{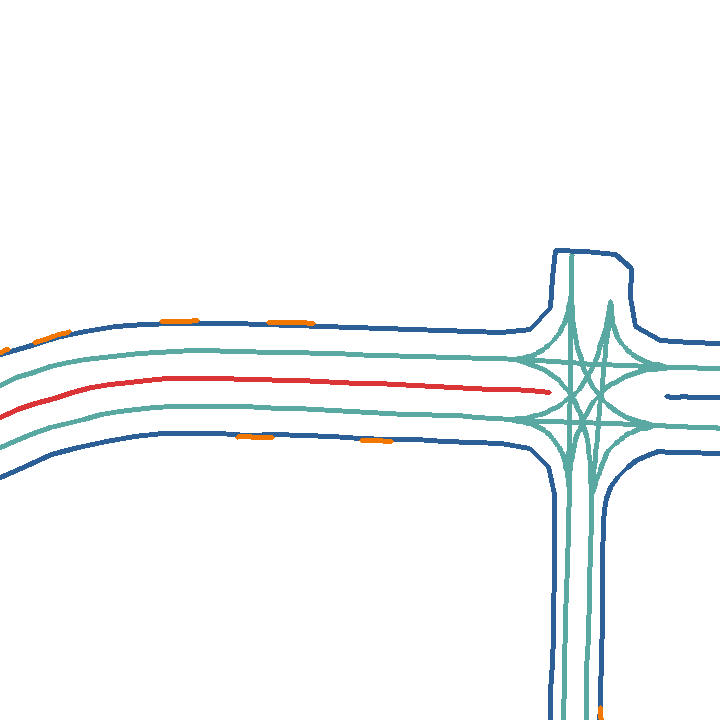}}
        \caption{Prior-Informed Prediction}
        \label{fig:}
    \end{subfigure}\hfill
    \begin{subfigure}[m]{0.24\textwidth}
        \frame{\includegraphics[width=\textwidth]{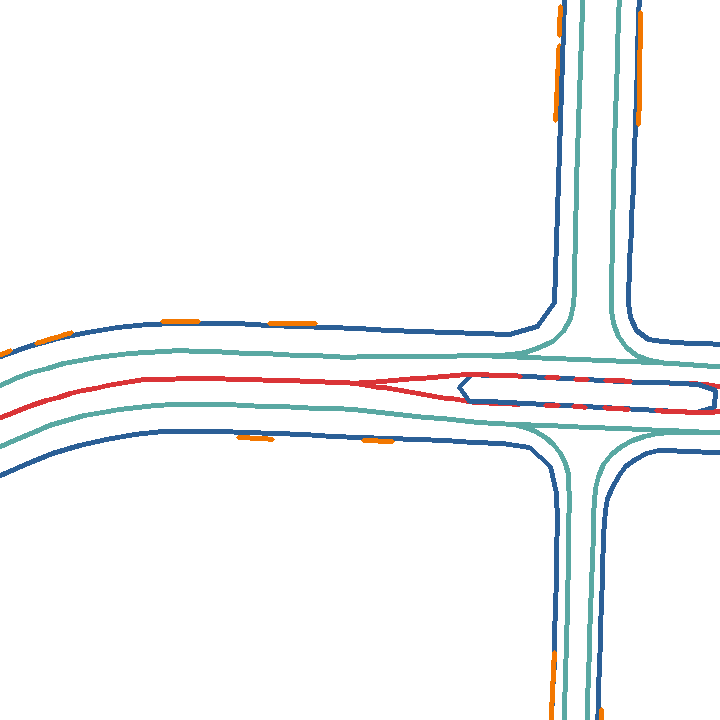}}
        \caption{Ground Truth}
        \label{fig:}
    \end{subfigure}

    \caption{Qualitative results handling real world changes. For minor real-world changes, e.g. driveway geometry (a--d) and curb geometry (e--h), a model trained with prior perturbations correctly predicts changes to many of the real-world features. However, for substantial changes in road layout, e.g. additional medians (i--l) and new road construction (m--p), the model fails to meaningfully deviate from the prior to account for the new intersection geometry. Note that the topdown map is centered on the vehicle in all figures.}
    \label{fig:qualitative_results}
\end{figure*}

\subsection{Real-World Change Examples}
\label{sec:data_mining}
To generate a large set of real world map change examples, we leverage an internal map database to generate a region change diff between a 2020 version of our internal HD map, and that same map from 2023. 
We only include changes to the map significant enough that they required a recollect of data in the area and subsequent changes to the HD semantic map.
We then mine for 30 second scenes collected in the second half of 2023 which intersect these regions of change and occur after all computed map changes. 
To dump data for these scenes, we obtain the prior from the 2020 map version and the ground truth labels from the map version at data collection time in 2023, something not available from public change detection datasets \cite{lambert_trust_2021} to the best of our knowledge.
With this approach, every scene mined contains some change in the HD map of varying severity. 

\subsection{Experiment Setup}
We perform all experiments with a BEV backbone similar to \cite{liu_bevfusion_2022} with pre-trained LiDAR and camera BEV feature extractors which are trained for general perception tasks.
We train each model for 75K steps on 32x Nvidia A100s on a large internal dataset collected from Houston, TX and Mountain View, CA
($>$13.7K scenes, $>$822K unique training frames), with a 90m square field of view centered on the vehicle and four different feature types: lane centers, lane dividers, road boundaries, and driveways. 
For evaluation, we utilize two major types of datasets, a synthetic evaluation set and a real world change dataset. 
First, for the synthetic evaluation set, we use a geographically split holdout test dataset ($>$3.3K scenes, 198K unique test frames) with low levels of synthetic prior noise utilizing all presented mutations (0.1m Std. Dev. for continuous warping mutations, and 0.1 probability for discrete mutations). 
Then, for real world change data, we mine a holdout test set (1240 scenes, 74k unique test frames) of scenes which have undergone real world HD map change, and provide an outdated map prior from multiple years of aggregated map changes in the geospatial map database as described in \cref{sec:data_mining}.

The value of utilizing our internal dataset is multifold. 
For one, it significantly reduces the risk of geospatial overfitting of map detection transformers \cite{lilja_localization_2023} by holding out large geospatial regions for eval exclusively during the training and evaluation split.
In addition, our dataset is significantly larger than the majority of open datasets (e.g. \cite{lambert_trust_2021}), and is thus able to leverage the scaling behavior of map prediction models identified in \cite{lilja_localization_2023} to further mitigate the noise caused by overfitting.
Finally, we provide real world pairs of outdated and up to date maps, something not available in \cite{lambert_trust_2021} as pointed out in \cite{sun_mind_2023}.
As for metrics, we compute mean Average Precision (mAP) of predicted polylines using the same Chamfer distance based metrics and thresholds used in \cite{liao_maptr_2023}.

\subsection{Experimental Design}

Due to the combinatorial hyperparameter space induced by so many different parameters, we first train a baseline model with a small amount of noise for each prior mutation, then tune each mutation parameter independently.
Qualitative results for the baseline low noise model are shown in \cref{fig:qualitative_results}.

For the parameter search (\cref{tab:discrete-parameter-search}, \cref{tab:continuous-param-search}), we start with all mutations except Perlin warp enabled with a low noise level (``Low All Noise" in the table), where this low noise level is identical to the synthetic perturbation distribution (0.1m Std. Dev for continuous mutations, 0.1 probability of discrete mutations).
We then test a number of increased levels of noise for their performance against the synthetic evaluation dataset as well as the real world evaluation dataset.

\begin{table}[h]
\begin{tabularx}{\linewidth}{c c >{\columncolor{lightgray}} X X}
\toprule
\multicolumn{1}{c}{\multirow{2}{*}{\textbf{Mutation}}}
& \multicolumn{1}{c}{\multirow{2}{*}{\textbf{Probability}}}
&\multicolumn{2}{c}{\textbf{mAP }}
\\
\multicolumn{2}{c}{} 
& Sim & Real
\\\midrule
Baseline (No Noise)  &   0.0 &  0.8980         &  0.8239  \\
Low All Noise &    0.1 &   \textbf{0.9934}       &   \textbf{0.8571}   \\
\midrule
\multirow{5}{*}{Drop Features} & 0.2   & \textbf{0.9917} &   \textbf{0.8564}   \\
& 0.4   & 0.9834 &   0.8435   \\
& 0.6   & 0.9677 &   0.8313   \\
& 0.8   & 0.6012 &   0.6074   \\
& 1.0   & 0.6580  &  0.6952    \\
\midrule
\multirow{3}{*}{Duplicate Features}  & 0.2 &  \textbf{0.9920} &   0.8535  \\
& 0.3  & 0.9892 & 0.8570 \\
& 0.5 & 0.9917 & \textbf{0.8604} \\
\midrule
\multirow{3}{*}{Wrong Class}    &0.2 &  0.9929   &   \textbf{0.8564}  \\
& 0.3  & \textbf{0.9932} & 0.8555 \\
& 0.5 & 0.9901 & 0.8542 \\

\bottomrule
\end{tabularx}
\caption{Discrete Mutation Parameter Search.}
\label{tab:discrete-parameter-search}
\end{table}

\begin{table}[h]
\begin{tabularx}{\linewidth}{c c >{\columncolor{lightgray}} X X}
\toprule
\multicolumn{1}{c}{\multirow{2}{*}{\textbf{Mutation}}}
& \multicolumn{1}{c}{\multirow{2}{*}{\textbf{Std. Dev}}}
&\multicolumn{2}{c}{\textbf{mAP }}
\\
\multicolumn{2}{c}{} 
& Sim & Real
\\\midrule
Baseline (No Noise)  &   0.0 m &  0.8980         &   0.8239   \\
Low All Noise &    0.1 m &   \textbf{0.9934}       &  \textbf{0.8571}   \\
\midrule
\multirow{3}{*}{Perturb Control Points} & 0.5 m  & \textbf{0.9929}  &   \textbf{0.8640}   \\
& 1.0 m  & 0.9843 &    0.8512  \\
& 2.0 m  & 0.9531 &   0.8476   \\
\midrule
\multirow{3}{*}{Shift Features}  & 0.5 m &     \textbf{0.9794}        &  \textbf{0.8612}   \\
& 1.0 m & 0.9373 & 0.8461 \\
& 2.0 m & 0.8781 & 0.8108 \\
\midrule
\multirow{3}{*}{Localization Noise}    & 0.5 m, $0.5^{\circ}$ & \textbf{0.9936}    &   \textbf{0.8648}  \\
& 1.0 m, $1.0^{\circ}$  & 0.9913 & 0.8588 \\
& 2.0 m, $2.0^{\circ}$ & 0.9911 & 0.8604 \\
\midrule
\multirow{3}{*}{Perlin Warp}    & 0.5 m &   \textbf{0.9854}  &   0.8606  \\
& 1.0 m & 0.9711 & \textbf{0.8623} \\
& 2.0 m & 0.9357 & 0.8407 \\

\bottomrule
\end{tabularx}
\caption{Continuous Warp Mutation Parameter Search.}
\label{tab:continuous-param-search}
\end{table}

%% file: sec/5_discussion.tex
\section{Discussion}
We note a number of interesting observations from our experimental results.
First, consistent with \cite{sun_mind_2023}, we note that training with no prior noise at all results in a ``pass-through'' model which learns to replicate the prior map without modifications.
Since changed map features usually comprise a small percentage of the features in any given frame of data, we see that this pass-through model which is ignoring the sensor BEV information is capable of achieving a $0.824$ mAP on the real world map changes regardless. 

More substantially, we note a consistent sim2real gap between the simulated, low noise evaluation prior and the real world map change dataset, where the model has learned to effectively denoise the evaluation prior but is not sufficiently general to smoothly transfer to real change detection. 
Error is correlated between the simulated and real prior corruption, but our simulated evaluation is insufficient to model the complexities of real world changes.
The qualitative results in \cref{fig:qualitative_results} reinforce this observation, where only the simplest real-world changes are accurately predicted by the model, which reverts to the prior when the changes become too complex.

Somewhat surprising is that increasing prior dropout noise primarily serves to degrade model performance on real world map changes as its noise is increased, rather than cleanly trading off between real world sensor and prior contributions.
Instead, we see that performance slowly degrades, until a bifurcation in response behavior when having too degraded of a prior causes the model to perform even worse than it does when trained with no prior and then tested with prior (Drop Features, p=1.0), which is completely out of distribution.
Similarly interesting is that increased likelihood of feature duplication is somewhat more helpful than any other discrete mutation, with performance increasing with increased mutation likelihood for the values we tested on. 

We see a slightly different story with continuous warping mutations, which all improve on real world performance with an increased level of noise from the baseline low noise level.
Past that, however, they also see degradation behavior as perturbations get more exaggerated at higher noise levels, similar to that of the discrete features.
This is likely due to the true distribution of map changes having a similar level of average displacement, where higher noise levels are unreasonable in nominal real world change scenarios (e.g. redoing a curb or driveway).

%% file: sec/6_conclusions.tex
\section{Conclusions}
\label{sec:conclusions}
Robustness to real-world changes is critical for any map-based autonomous vehicle system.
We confirm the conclusions of \cite{sun_mind_2023} in that prior maps are much better than no prior and that we need some noise in the prior to learn something more useful than a pass-through function. 

However, we are able to expand on those results by observing that too corrupted or weak of a prior can actually harm performance of the model more than omitting the prior entirely.
Most importantly, we show that there exists a considerable sim2real gap between real world change detection performance and performance on simulated prior noise.
We observe through large-scale experiments that prior mutations are sufficient to capture only the simplest of real-world changes. 
We hope the results presented in this paper motivate future work in this area to address the sim2real gap for HD map prediction with prior.